\documentclass[conference]{IEEEtran}
\IEEEoverridecommandlockouts

\usepackage{fancyhdr}

\fancypagestyle{firstpage}{%
  \fancyhead{}
  
  \fancyfoot[C]{Accepted for Proceedings of WCCI 2026, Maastricht, Netherlands. Copyright IEEE}
}
\usepackage{cite}
\usepackage{float}
\usepackage{amsmath,amssymb,amsfonts}
\usepackage[ruled,vlined]{algorithm2e}
\usepackage{graphicx}
\usepackage{gensymb}
\usepackage{textcomp}
\usepackage{xcolor}
\usepackage{steinmetz}
\usepackage{makecell}
\usepackage{tabularx}
\usepackage{balance}
\usepackage{subcaption}
\def\BibTeX{{\rm B\kern-.05em{\sc i\kern-.025em b}\kern-.08em
    T\kern-.1667em\lower.7ex\hbox{E}\kern-.125emX}}
\begin{document}

\title{The Role of Grid Cells in Reducing Spatial Aliasing in Hippocampal Place Representations \thanks{\IEEEauthorrefmark{1}These authors contributed equally to this work.}}

\author{
\IEEEauthorblockN{Alexander Johnson\IEEEauthorrefmark{1}}
\IEEEauthorblockA{\textit{Dept. of Electrical} \\
\textit{and Computer Engineering}\\
\textit{University of Cincinnati}\\
Cincinnati, USA \\
johns9a4@mail.uc.edu}
\ \\
\and 
\IEEEauthorblockN{Obadah Ghizawi\IEEEauthorrefmark{1}}
\IEEEauthorblockA{\textit{Dept. of Electrical} \\
\textit{and Computer Engineering}\\
\textit{University of Cincinnati}\\
Cincinnati, USA \\
ghizawon@mail.uc.edu}
\and
\IEEEauthorblockN{Ali A. Minai, {\it Senior Member IEEE}}
\IEEEauthorblockA{\textit{Dept. of Electrical} \\
\textit{and Computer Engineering}\\
\textit{University of Cincinnati}\\
Cincinnati, USA \\
minaiaa@ucmail.uc.edu}
}

\maketitle

\thispagestyle{firstpage}


\begin{abstract}
Spatial aliasing occurs when two or more distinct locations produce highly similar place-cell representations, primarily due to environmental symmetry or repetitive structures. This issue is most pronounced when place representations are constructed solely from boundary vector cell (BVC) inputs, because symmetric or repetitive structures can yield indistinguishable sensory patterns across multiple locations in an environment. This work introduces grid cell signals to mitigate spatial aliasing in such settings. Because grid cells contribute periodic, internally generated spatial signals that vary independently of environmental geometry, they play a key role in disambiguating perceptually identical locations. We integrate multiple modules of analytically constructed grid cells with BVC-driven place cells and show that this leads to a 94--99\% reduction in spatial aliasing relative to a BVC-only baseline across three environments: an open environment without obstacles; an environment with a cross-shaped central obstacle creating high visual symmetry; and a maze environment. The greatest improvement occurs in the environment with the highest visual symmetry. These results indicate that grid cells provide information complementary to boundary-based inputs, yielding more reliable place representations in geometrically ambiguous environments.

\end{abstract}

\begin{IEEEkeywords}
grid cells, place cells, spatial navigation, spatial cognition
\end{IEEEkeywords}


\section{Introduction}
\label{sec:intro}
Biologically-inspired computational models of spatial cognition aim to replicate the navigational abilities observed in animals. These models are often derived from the hippocampal region of the brain, which is known to be central for spatial mapping and navigation in mammals \cite{moser2008place}. Place cells are specialized neurons in the hippocampus that exhibit spatially localized regions of activity, known as place fields \cite{okeefe1971hippocampus,o1978hippocampus, best2001spatial}. Place field firing patterns have been shown to depend on distal visual cues \cite{muller1987effects}, and especially cues about the distance and location of walls and other objects mediated through boundary vector cells (BVCs) \cite{hartley2000modeling, o1996geometric}. However, maps built using only these cues are susceptible to spatial aliasing, in which the same place cells are activated at multiple spatial locations due to visual symmetries in the environment 
\cite{GerstenslagerDukenbaevMinai2025}. 

Grid cells are neurons in the medial entorhinal cortex (MEC), whose periodic spatial firing fields at different orientations and spatial frequencies form triangular lattices that cover any environment visited by an animal~\cite{bush2014what, hafting2005microstructure}. According to Bush et al. \cite{bush2014what}, grid cells provide a context-dependent spatial metric for path integration and vector navigation, and these complementary signals interact with place cells to support reliable coding of large environments. In this paper, we show that one way that grid cells improve the accuracy of spatial representations is by reducing aliasing. They do this by introducing a secondary source of information, thereby disambiguating visually similar locations.

\textit{Note: The simulator used for this work is part of a larger hippocampal system simulation platform being developed for use with the Webots environment. It will be shared publicly once it is complete and described fully in a future report.}

\section{Background}
\label{sec:bio_background}
\subsection{Place Cells}
The first report of place-specific activity in hippocampal cells was by O'Keefe et al. in 1971 \cite{okeefe1971hippocampus}. Place cells (PCs) are primarily located in the CA3 and CA1 regions of the hippocampus. PCs exhibit location-specific and boundary-aware firing, meaning that a given cell fires only when the animal is in a particular part of the environment (i.e., the place field) and maintains a tunable distance (and direction) from boundaries. For example, when a place cell is recorded in a number of rectangular environments, the peak response location of the PC often maintains a fixed distance from the two nearest walls \cite{burgess1996neuronal}. A PC's firing rate is modeled as a function of the animal's location through the thresholded sum of firing rates of several inputs (traditionally, BVCs) \cite{hartley2000modeling}. Place cells are a critical component of the system that enable mammals to represent their location within a given environment.

\subsection{Head Direction Cells}
Head direction cells (HDCs) were first recorded and characterized by Taube et al. in 1990. These cells are located in the postsubiculum (Brodmann area 48), a key region in the hippocampal formation. Each cell fires rapidly only when the animal's head is pointing in a restricted range of angles, with the maximum activation occurring in approximately the middle of this range. This angle is called the {\it preferred direction} of the cell \cite{taube1990headI}. Together, activated head direction cells provide a population-coded signal indicating the way an animal is facing at any given time. Head direction cells are an important component of the hippocampus's navigational circuitry.

\subsection{Boundary Vector Cells}
Boundary vector cells (BVCs) were first predicted to exist through computational models using hippocampal place cells in 1996 \cite{o1996geometric}, and later in 2000 \cite{hartley2000modeling, burgess2000predictions}. BVCs were then confirmed as a cell type in 2009 by Lever et al. \cite{lever2009boundary}. A BVC fires whenever an environmental boundary intersects a receptive field located at a specific distance (i.e., preferred distance) from an animal in a specific allocentric direction. The firing of a BVC only depends on the animal's location, rather than the animal's heading \cite{lever2009boundary}. BVCs are a key input to place cells (PCs), as the firing of a PC is a thresholded sum of the firing of the BVCs connected (via synapses) to it \cite{hartley2000modeling}. This means that BVCs provide critical information about surrounding boundaries and their directions to PCs, allowing PCs to activate at specific BVC configurations and, thus, for each PC to have an independent role in the navigational system.

\subsection{Grid Cells}
Entorhinal grid firing patterns were first discovered by Hafting et al. in 2005 \cite{hafting2005microstructure}. The dorsocaudal medial entorhinal cortex (dMEC) contains a topographically organized neural representation of space, with the key units being grid cells. A grid cell exhibits multiple spatial firing fields that form a periodic hexagonal lattice across the environment. It has been shown that grids of neighboring cells differ in terms of their vertex locations (i.e., their phases), with the spacing and size of individual fields increasing from dorsal to ventral dMEC \cite{hafting2005microstructure}. Grid cells are a core component of the MEC, supporting path integration and enabling navigation even in the absence of external landmarks.

\subsubsection{Grid Cell Modeling}
A variety of computational models have been proposed to explain GC firing patterns. These can be broadly categorized as either \textit{dynamical} or \textit{learning-based} models. Dynamical models propose that grid patterns emerge from the temporal evolution of neural activity. For example, oscillatory inference models \cite{burgess2007oscillatory, hasselmo2007grid} rely on interference patterns of theta-frequency oscillations, which are modulated by velocity. Another dynamical approach is to use continuous attractor networks (CANs) \cite{burak2009accurate}, where grid patterns emerge as stable activity ``bumps'' that shift in response to velocity input. Alternatively, learning-based approaches have demonstrated that grid-like representations can naturally emerge in deep neural networks trained on vector-based navigation \cite{banino2018vector}. 

As this work focuses on the functional utility of GCs rather than their biological origin, we adopt a direct analytical approach to their implementation. By explicitly constructing grid-like firing patterns using cosine interference, we achieve precise control over grid scale and orientation while maximizing computational efficiency. The goal is not to reproduce the physiology of grid cells, but to leverage their functional properties—namely, periodic, metrically consistent spatial information—to reduce place-field aliasing in complex, symmetric environments. 

\subsection{Spatial Aliasing}
Spatial aliasing is a phenomenon where similar place representations arise in visually similar locations. This problem is especially serious in models in which place cell activity is driven mainly by sensory inputs, such as BVC activity. While some spatial aliasing has been observed in animals, and even humans may occasionally be confused by similar-looking locations, hippocampal representations of different locations are usually quite distinct and can be used to localize the animal's position in experimental setups \cite{davidson2009replay,carr:2011replay,dragoi2011preplay}. Recent work in our group has shown that, when available, the 3-dimensional structure of the environment may help mitigate the problem \cite{GerstenslagerDukenbaevMinai2025}

In this paper, we present a more robust and universally applicable solution to spatial aliasing using grid cells.

\begin{figure}[t]
    \centering
    \includegraphics[width=0.60\columnwidth]{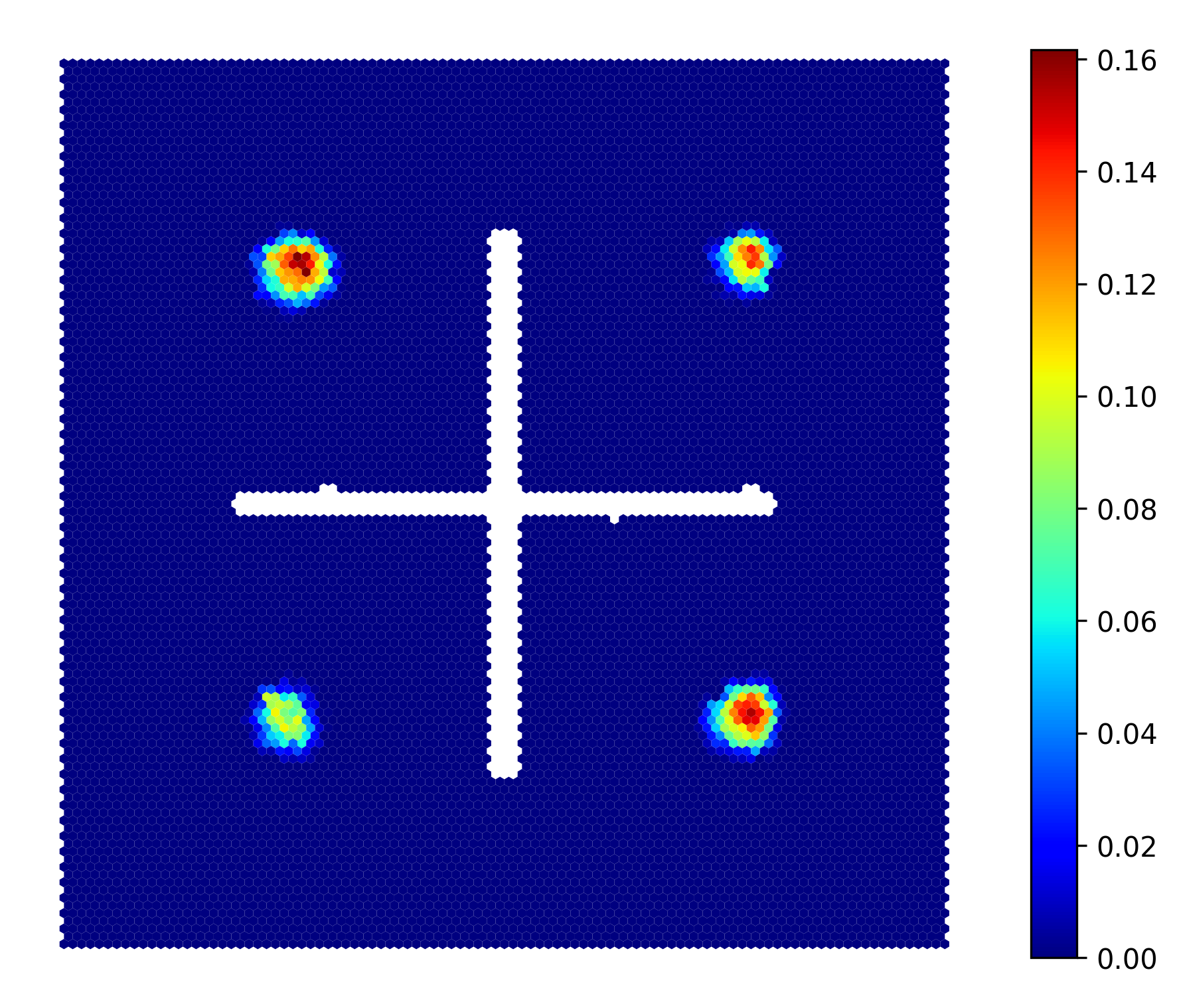}
    \caption{Example of spatial aliasing in a single place cell. In the Cross environment, this neuron exhibits multiple firing-field peaks (four distinct activation maxima), indicating ambiguous spatial encoding where separated locations elicit similar place cell responses.}
    \label{fig:aliased_place_cell_four_peaks}
\end{figure}

\section{Model Architecture}
\label{sec:model}
The model consists of four core layers, each modeled after its biological counterpart: Head Direction Cell (HDC), Boundary Vector Cell (BVC), Grid Cell (GC), and Place Cell (PC) layers. The HDC layer encodes the agent's heading, with each cell tuned to a specific allocentric direction. The BVC layer encodes obstacle and boundary information, with each cell exhibiting preferred angular and distance tuning. The GC layer serves as an internal positional signal for the agent and is organized into discrete modules of grid cell populations. All modules share the same spatial scale but differ in phase offset. The PC layer serves as the model's core spatial representation and receives weighted input from both the BVC and GC layers.

The HDC, BVC, and PC layers are adapted from previous models~\cite{alabi2020oneshot,alabi2023}. 

The main contribution of this work is the addition of the GC layer, which provides a metric spatial coordinate system that complements the boundary-based representations and enables stable place field formation in environments with geometric ambiguity.

\begin{figure}[t]
    \centering
    \includegraphics[width=\columnwidth]{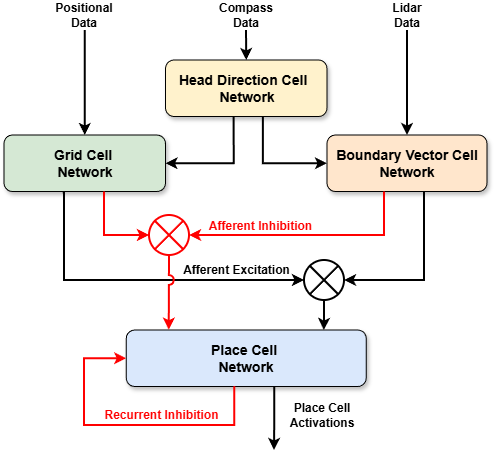}
    \caption{Model architecture for spatial representation. Black arrows indicate excitatory pathways, while red arrows indicate inhibitory pathways. Positional input drives the grid cell network, compass input drives head direction cells, and lidar input drives boundary vector cells (BVCs). Grid cell and BVC signals provide afferent excitation to the place cell network, while afferent and recurrent inhibition regulate competition and sparsity. Head direction signals provide a shared orientation reference between grid and boundary representations.}
    \label{fig:model_architecture}
\end{figure}

\subsection{Notation \& Overview}
For each of the four cell types---Head Direction (\textit{h}), Boundary Vector (\textit{b}), Grid (\textit{g}), and Place (\textit{p})---we use the superscript symbols \textit{h}, \textit{b}, \textit{g}, and \textit{p}, respectively. The time-varying firing rate of the $i^{\text{th}}$ cell of type $k$ is denoted by $v_i^k(t)$ (or simply $v_i^k$). Similarly, the time-varying synaptic weight from the $j^{\text{th}}$ cell of type $l$ to the $i^{\text{th}}$ cell of type $k$ is denoted as $W_{ij}^{kl}(t)$ (or $W_{ij}^{kl}$ when time indices are omitted). These definitions apply throughout all network modules and environments unless otherwise specified.

\subsection{Head Direction Cell Layer}
\label{sec:hd}

The Head Direction Cell (HDC) layer provides a global orientation signal to the model, where each cell has a preferred allocentric direction. For simplicity, this model uses $N_h = 8$ cells, whose preferred directions are fixed at $0^\circ, 45^\circ, \ldots, 315^\circ$. 

Each HDC $i$ has a preferred direction $\theta_i^h$ and activates according to:
\begin{equation}
v_i^h(t) = \cos\!\big(\theta(t) - \theta_i^h\big),
\label{eq:hd}
\end{equation}
where $\theta(t)$ is the agent's current global heading. This cosine tuning, derived from the dot product $[\cos\theta(t), \sin\theta(t)] \cdot [\cos\theta_i^h, \sin\theta_i^h]$, produces maximal response when the agent faces the cell's preferred direction, decreasing smoothly with angular deviation. This formulation is adapted from Erdem and Hasselmo~\cite{erdemHasselmo2012}.

\subsection{Boundary Vector Cell Layer}
\label{sec:bvc}

The Boundary Vector Cell (BVC) layer provides geometric spatial information by encoding the agent's relationship to environmental boundaries. This implementation follows the boundary vector formulation of Barry et al.~\cite{barry2006boundary}.

Each BVC $i$ is characterized by a preferred boundary displacement, specified by a radial distance $d_i$ and an allocentric bearing $\phi_i$ relative to the agent. The layer receives sensory input from LiDAR in the form of per-beam range measurements
$\mathbf{r}(t) = \{r_1, r_2, \ldots, r_{n_{\text{res}}}\}$ 
at fixed allocentric angles 
$\{\theta_1, \theta_2, \ldots, \theta_{n_{\text{res}}}\}$,
where $n_{\text{res}}$ denotes the angular resolution of the sensor (720 in this work).

Each BVC integrates evidence across all sensor beams by combining radial and angular Gaussian tuning functions. With tuning widths $\sigma_r$ (distance) and $\sigma_\theta$ (angle), the firing rate of BVC $i$ is:
\begin{equation}
    v_i^b(t) 
    \;=\; \frac{1}{N_{\text{BVC}}} \sum_{j=1}^{n_{\mathrm{res}}}
    \Biggl(\,
        \frac{\exp\!\bigl[-\tfrac{(r_j - d_i)^2}{2\,\sigma_r^2}\bigr]}{\sqrt{2\pi}\,\sigma_r}
        \;\times\;
        \frac{\exp\!\bigl[-\tfrac{(\theta_j - \phi_i)^2}{2\,\sigma_\theta^2}\bigr]}{\sqrt{2\pi}\,\sigma_\theta}
    \Biggr),
\label{eq:bvc}
\end{equation}
where $N_{\text{BVC}}$ is the total number of boundary vector cells. The normalization by $N_{\text{BVC}}$ ensures activations remain in a consistent range for different BVC population sizes. Preferred displacements $(d_i, \phi_i)$ are uniformly distributed to tile the boundary coding space: distances $d_i$ span $[0, r_{\max}]$ (the maximum LiDAR range), while bearings $\phi_i$ cover $[0, 2\pi)$. This arrangement ensures comprehensive coverage of potential boundary configurations surrounding the agent.

\subsection{Grid Cell Layer}
\label{sec:gc}

The introduction of grid cells into the model was motivated by the desire to address a key limitation of prior models: severe place-field aliasing in environments with symmetry or repeated geometry. Grid cells provide an internally generated representation of space by combining their hexagonal firing patterns.  This periodic structure provides a unique spatial signature at every location. When observing population activity within a grid cell module—where cells share a common scale and orientation but differ in spatial phase—the combined pattern becomes highly distinctive. Importantly, this spatial code is not affected by environmental geometry in the same way as boundary-based input. Even in the presence of repeated geometry or symmetrical layouts, the relative phase of grid cell activity differs across locations, providing the information needed to disambiguate them. 

This work uses an analytical approach to grid cell modeling: we directly construct the hexagonal periodic structure mathematically rather than simulating underlying neural mechanisms~\cite{blair2007scale}. This provides computational efficiency and precise control over spatial properties while maintaining biological plausibility through post-processing and obstacle-aware masking. The model implements $M = 8$ grid cell modules aligned with head direction cells at orientations $\alpha_m \in \{0^\circ, 45^\circ, \ldots, 315^\circ\}$, each containing $N_m = 50$ cells sharing a common spatial scale $\lambda$ but differing in spatial phase offset $\boldsymbol{\phi}_j$.

\subsubsection{Analytical Grid Cell Construction}

For a grid cell with orientation $\alpha$, spatial scale $\lambda$, and phase offset $\boldsymbol{\phi} = (\phi_x, \phi_y)$, we define the spatial frequency $\omega = 2\pi/\lambda$. The position $\mathbf{x} = (x, y)$ is rotated into the grid cell's reference frame and translated by its phase offset:
\begin{equation}
\begin{pmatrix} r_x \\ r_y \end{pmatrix} 
= \begin{pmatrix} \cos\alpha & \sin\alpha \\ -\sin\alpha & \cos\alpha \end{pmatrix}
\begin{pmatrix} x - \phi_x \\ y - \phi_y \end{pmatrix}.
\label{eq:rotation}
\end{equation}

The canonical grid pattern is constructed as the mean of three cosine gratings oriented $60^\circ$ apart:
\begin{equation}
\begin{split}
G(\mathbf{x}) &= \frac{1}{3}\Big[\cos(\omega r_x) \\
&\quad + \cos\Big(\omega \big(\tfrac{r_x}{2} + \tfrac{\sqrt{3}}{2} r_y\big)\Big) \\
&\quad + \cos\Big(\omega \big(\tfrac{r_x}{2} - \tfrac{\sqrt{3}}{2} r_y\big)\Big)\Big].
\end{split}
\label{eq:canonical-grid}
\end{equation}

To achieve the sharply tuned firing fields observed in biological recordings~\cite{hafting2005microstructure}, three post-processing steps are applied. First, a power-law transformation adjusts the activation profile:
\begin{equation}
G'(\mathbf{x}) = \text{sgn}(G) \cdot |G|^{1/\beta},
\label{eq:grid-spread}
\end{equation}
where $\beta \in [1.2, 1.8]$ controls field sharpness. Second, per-cell min-max normalization ensures consistent dynamic range. Third, soft thresholding creates sparse activations:
\begin{equation}
v^g(\mathbf{x}) = \max\Big(0, \frac{G''(\mathbf{x}) - \theta_{\text{gc}}}{1 - \theta_{\text{gc}}}\Big),
\label{eq:grid-threshold}
\end{equation}
where $\theta_{\text{gc}}$ controls sparsity.

\subsubsection{Obstacle-Aware Masking}

Analytical construction based on position alone produces activations that extend across walls, contradicting biological observations where grid cells fragment at barriers while maintaining phase coherence~\cite{derdikman2009fragmentation}. We address this through a spatial masking procedure applied independently to each grid cell.

The procedure identifies connected activation components (``blobs'') and determines which intersect obstacles. For blobs where obstacles cover $\ge 20\%$ of their diameter, we apply obstacle masks to split or suppress the activation. When a blob splits into multiple components, only the largest is retained. Blobs that remain connected after initial masking are processed with dilated obstacle boundaries to force fragmentation. Finally, Gaussian smoothing ($\sigma \approx 0.5$) restores biologically plausible curved edges that taper near walls rather than exhibiting sharp cutoffs.

\begin{figure}[t]
    \centering

    \begin{subfigure}[b]{0.48\columnwidth}
        \centering
        \includegraphics[width=\linewidth]{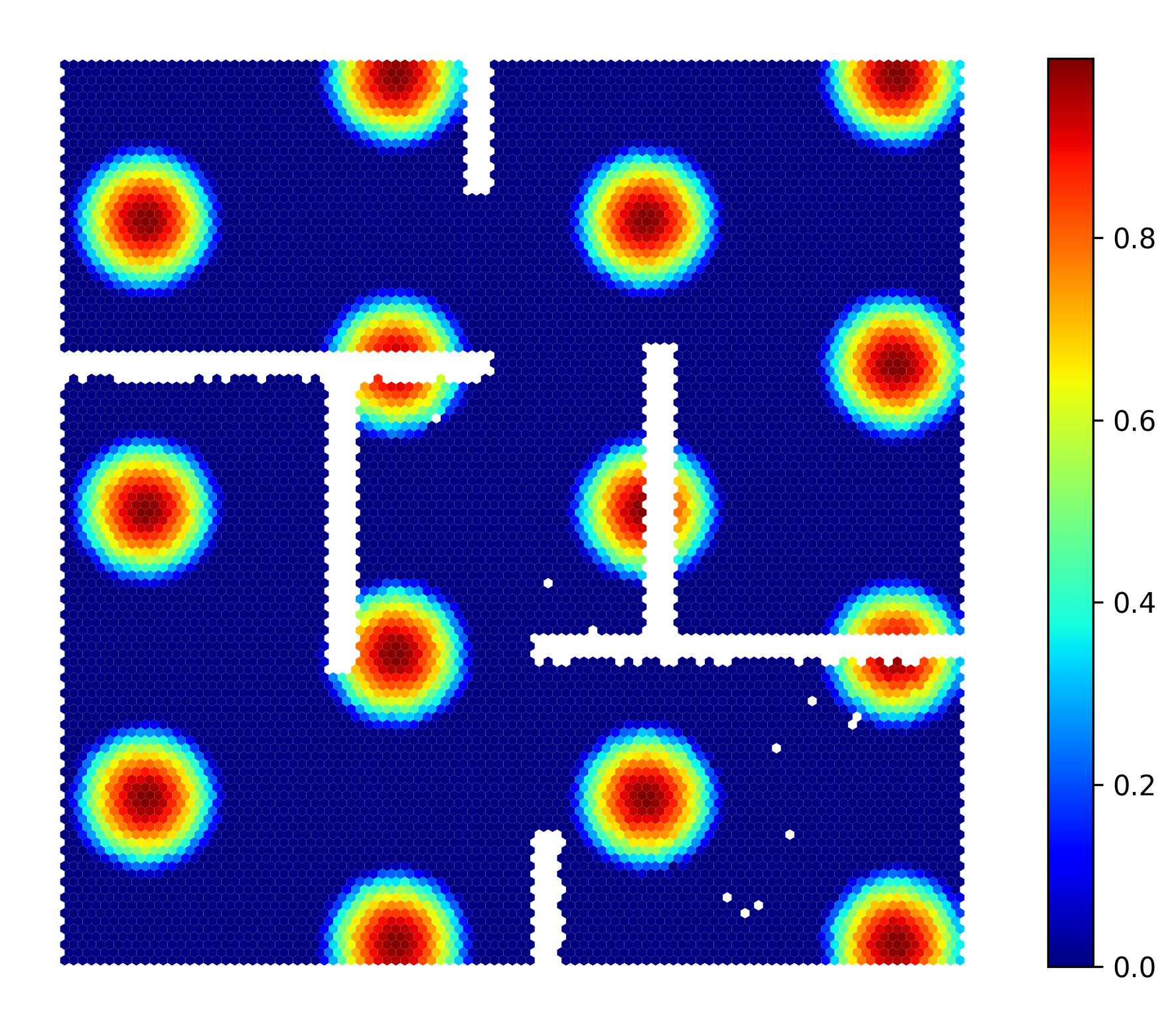}
        \caption{Raw (no masking)}
        \label{fig:boundary_masking_raw}
    \end{subfigure}
    \hfill
    \begin{subfigure}[b]{0.48\columnwidth}
        \centering
        \includegraphics[width=\linewidth]{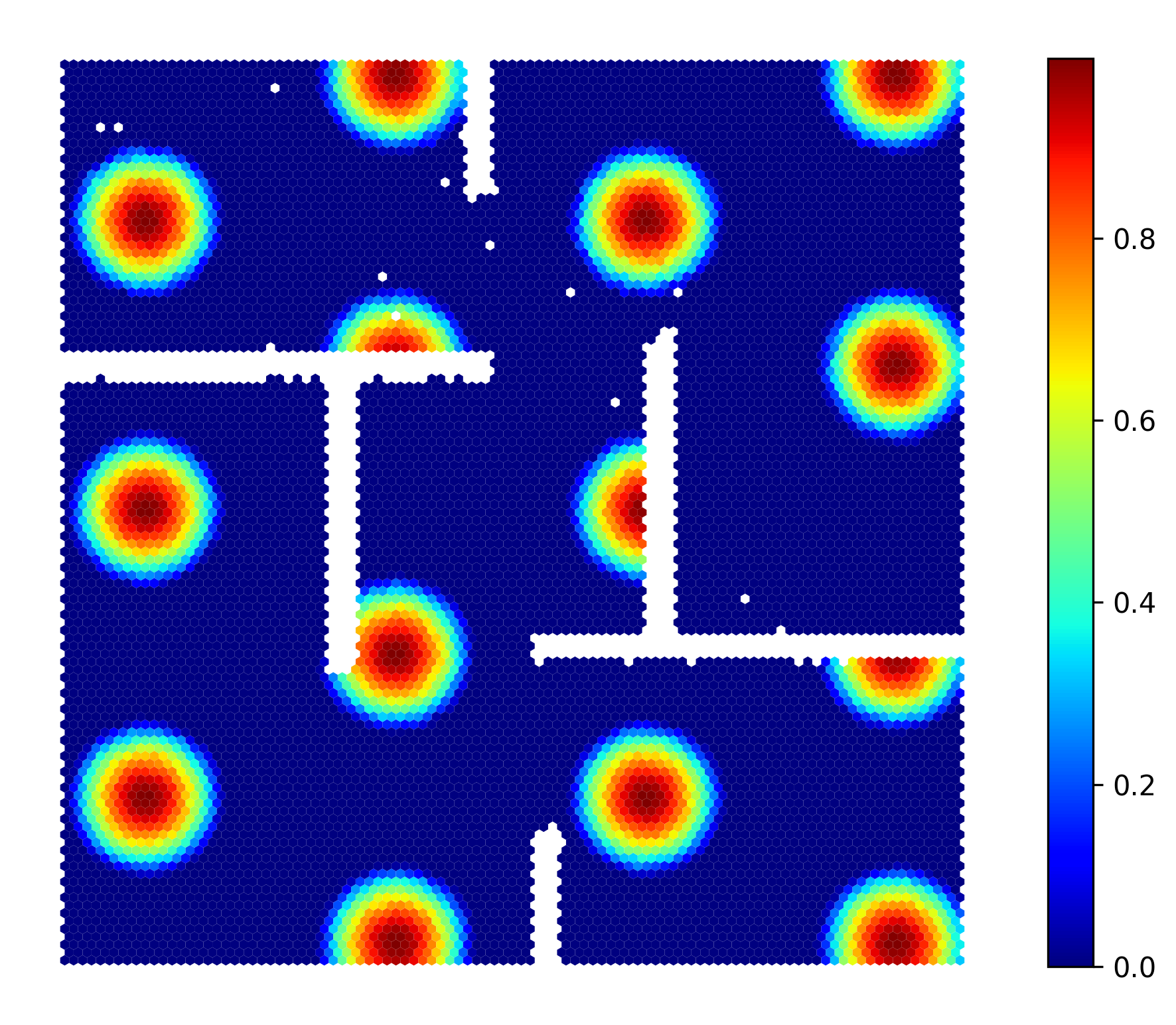}
        \caption{Masked (boundary-aware)}
        \label{fig:boundary_masking_masked}
    \end{subfigure}

    \caption{Boundary-aware masking for grid cell activations in an obstacle-filled environment. Without masking (left), activations extend across walls. With masking (right), activations fragment at barriers while preserving local phase coherence within each compartment.}
    \label{fig:boundary_masking_comparison}
\end{figure}

Figure~\ref{fig:boundary_masking_comparison} illustrates this effect: raw activations extend continuously across walls (left), while masked activations fragment at boundaries while preserving local phase coherence (right).

\subsection{Place Cell Layer}
\label{sec:pc}

The Place Cell (PC) layer forms the core spatial representation of the model. Place cells develop spatially localized receptive fields (place fields) through competitive learning, with each cell encoding a specific location in the environment. This layer integrates information from both BVC and GC layers: BVCs provide geometric constraints from environmental boundaries, while GCs contribute metric phase information that disambiguates perceptually similar locations. The relative contribution of these two input streams is controlled by a weighting parameter $\eta \in [0,1]$, where BVC inputs are weighted by $(1-\eta)$ and GC inputs by $\eta$.

\subsubsection{Membrane Dynamics and Activation}

The membrane potential $s_i^p$ of place cell $i$ evolves as a leaky integrator combining excitatory input from BVCs and GCs with global inhibition:
\begin{align}
\tau_p \,\frac{ds_i^p}{dt}
&= -s_i^p + E_i - I_i,
\label{eq:state-dynamics}
\end{align}
where the excitatory drive is
\begin{equation}
E_i = (1-\eta)\sum_{j} W^{pb}_{ij}\,v^b_j + \eta \sum_{k} W^{pg}_{ik}\,v^g_k,
\end{equation}
and the inhibitory drive is
\begin{equation}
I_i = (1-\eta)\,\Gamma^{pb}\!\sum_{j} v^b_j + \eta\,\Gamma^{pg}\!\sum_{k} v^g_k + \Gamma^{pp}\!\sum_{l} v^p_l.
\end{equation}
Here, $W^{pb}_{ij}$ and $W^{pg}_{ik}$ are synaptic weights from BVCs and GCs to place cell $i$; $v^b_j$, $v^g_k$, and $v^p_l$ are the firing rates of BVCs, GCs, and PCs respectively; and $\Gamma^{pb}$, $\Gamma^{pg}$, $\Gamma^{pp}$ are inhibitory gain parameters that control the strength of afferent (boundary and grid) and recurrent (place-to-place) inhibition. The firing rate is obtained through rectification and saturation:
\begin{equation}
v_i^p = \tanh\!\big([\psi\,s_i^p]_+\big),
\label{eq:nonlinearity}
\end{equation}
where $\psi$ is a gain factor and $[\cdot]_+ = \max(0,\cdot)$. This nonlinearity produces sparse, localized place fields through the combined effects of global inhibition and thresholding.

\subsubsection{Self-Organization via Competitive Learning}

Place fields emerge through competitive learning during exploration. Synaptic weights $W^{pb}_{ij}$ and $W^{pg}_{ik}$ are initialized sparsely with connection probabilities of approximately 0.25 and 0.30, respectively, ensuring each place cell receives input from a distinct subset of BVCs and GCs. During learning, weights are updated according to Oja's rule~\cite{oja1982simplified,alabi2020oneshot,alabi2023}:

\begin{align}
\tau_{w^{pb}} \frac{dW_{ij}^{pb}}{dt}
&= (1-\eta)\,v_i^p\!\left(v_j^b - \frac{1}{\alpha_{pb}}\,v_i^p\,W^{pb}_{ij}\right),
\label{eq:oja-bvc}
\\[4pt]
\tau_{w^{pg}} \frac{dW_{ik}^{pg}}{dt}
&= \eta\,v_i^p\!\left(v_k^g - \frac{1}{\alpha_{pg}}\,v_i^p\,W^{pg}_{ik}\right),
\label{eq:oja-grid}
\end{align}
where $\tau_{w^{pb}}$, $\tau_{w^{pg}}$ are learning time constants and $\alpha_{pb}$, $\alpha_{pg}$ are normalization factors that control the strength of weight decay. The first term in each equation strengthens synapses when pre- and postsynaptic cells are co-active (Hebbian learning), while the second term normalizes total synaptic input and enforces competition among place cells. Combined with global inhibition ($I_i$), this produces winner-take-all dynamics where each place cell specializes to represent a unique combination of BVC and GC activity patterns, forming a spatially localized place field.

\begin{figure}[t]
    \centering

    \begin{subfigure}[b]{0.32\columnwidth}
        \centering
        \includegraphics[width=\linewidth]{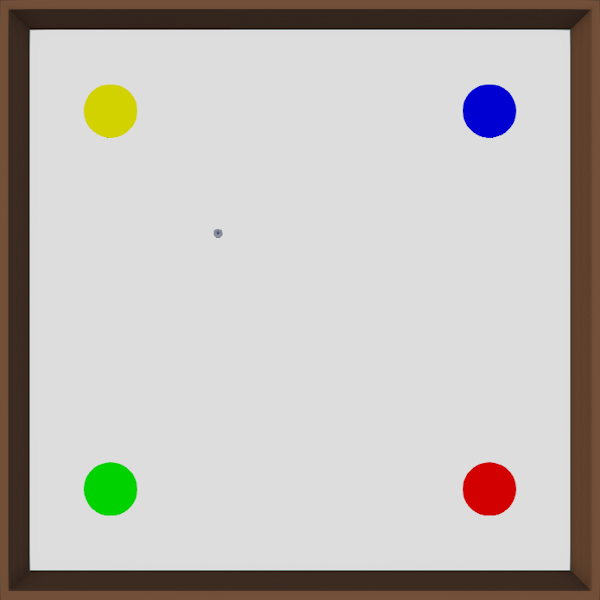}
        \caption{Open}
        \label{fig:env_open}
    \end{subfigure}
    \hfill
    \begin{subfigure}[b]{0.32\columnwidth}
        \centering
        \includegraphics[width=\linewidth]{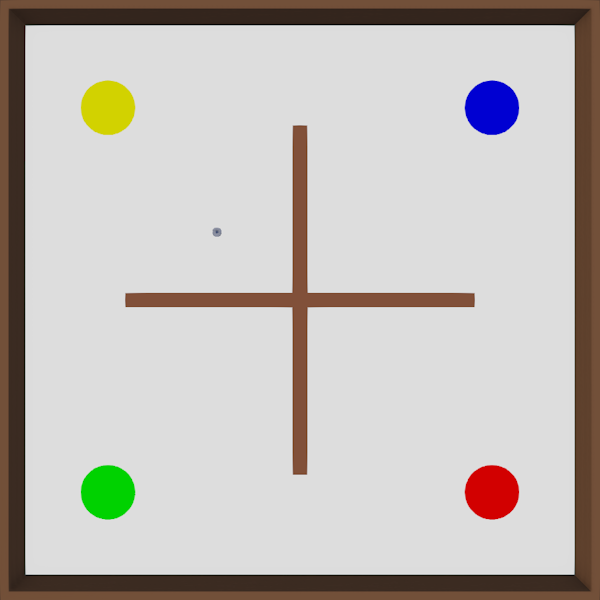}
        \caption{Cross}
        \label{fig:env_cross}
    \end{subfigure}
    \hfill
    \begin{subfigure}[b]{0.32\columnwidth}
        \centering
        \includegraphics[width=\linewidth]{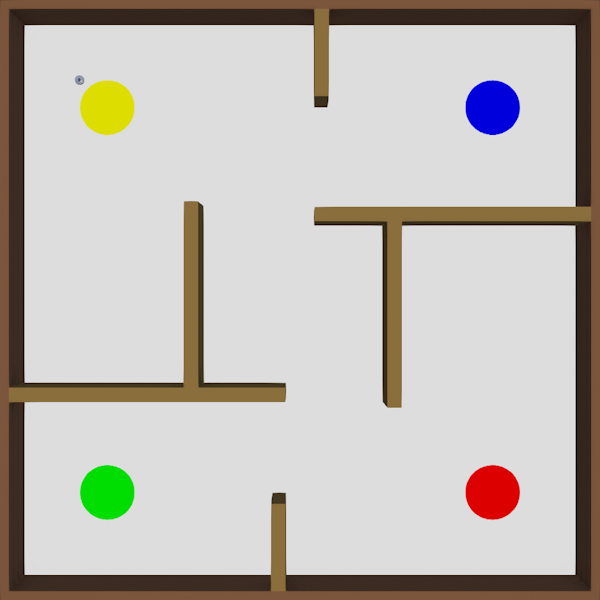}
        \caption{Maze}
        \label{fig:env_maze}
    \end{subfigure}

    \caption{Evaluation environments used in experiments: (a) Open, (b) Cross, and (c) Maze. All environments are 20m x 20m in size.}
    \label{fig:environments}
\end{figure}

\section{Experimental Setup}
\label{sec:exp_setup}
All experiments were conducted using an RTX 3090 GPU and a Ryzen 9 9900X CPU. The simulation software used to run the experiments was Webots R2025a. The agent used across the experiments was a Roomba bot with a compass for head-direction updates and a rangefinder with 720 beams covering 360 degrees, which provided information directly to the BVC layer. The rangefinder has a maximum distance of 25m, to account for the maximum distance that would need to be read in a 20\,m $\times$ 20\,m environment. All environments used are 20\,m $\times$ 20\,m in size. 

\subsection{Data Collection}
\label{sec:data_coll}
The agent explored three environments of varying complexities. The first is an open environment without obstacles, the second features cross-shaped walls that divide it into four regions, and the last is modeled as a maze, as shown in Fig. \ref{fig:environments}. To demonstrate the effects of grid cells in mitigating aliasing, 5 trials were conducted for each environment with grid cells, and another set of 5 trials was conducted without grid cells. The results across these trials were averaged for each case tested. Consequently, 10 trials were conducted for each environment, yielding a total of 30 trials. In a given trial, there were training and evaluation phases. During the training phase, the agent randomly explored the environment until it had covered at least 95 percent of it. This was calculated by dividing the environment into a square grid with a bin size of 0.5m and tracking the areas visited until the grid's coverage reached 95 percent. Bins that overlap with obstacles were not included in this calculation. During the evaluation phase, the agent explored until it reached 99 percent coverage, with a bin size of 0.2m, matching the bin size used to calculate the \textit{Mean Spatial Aliasing Index (MSAI)} metric (see below). 

\subsection{Model Parameters}
\label{sec:model_params}
\begin{center}
\captionof{table}{Neural Cell Populations Used in the Model}
\label{tab:cell_counts}

\begin{tabular}{l c}
\hline
\textbf{Cell Type} & \textbf{Number of Cells} \\
\hline
Place Cells & 1000 \\
Boundary Vector Cells (BVCs) & 400 \\
Head Direction Cells & 8 \\
Grid Cells & 400 \\
\hline
\end{tabular}
\end{center}

Table~\ref{tab:cell_counts} summarizes the number of cells used for each of the respective cell types within the experiments. 

\begin{center}
\captionof{table}{Boundary Vector Cell Parameters}
\label{tab:bvc_params}

\begin{tabular}{l c}
\hline
\textbf{Parameter} & \textbf{Value} \\
\hline
Radial Tuning Width ($\sigma_r$) & $1.0\,\mathrm{m}$ \\
Angular Tuning Width ($\sigma_\theta$) & $3.0^\circ$ \\
BVCs per Direction & 50 \\
Total Number of BVCs & 400 \\
\hline
\end{tabular}
\end{center}

Table~\ref{tab:bvc_params} summarizes the BVC parameters used within the experiments. Along each head direction, 50 BVCs were used. As summarized in Table~\ref{tab:cell_counts}, there are 8 head directions, meaning a total of 50 $\times$ 8 = 400 BVCs were used in the model. 

\begin{center}
\captionof{table}{Grid Cell Parameters}
\label{tab:grid_params}

\begin{tabular}{l c}
\hline
\textbf{Parameter} & \textbf{Value} \\
\hline
Number of Modules & 8 \\
Cells per Module & 50 \\
Total Number of Grid Cells & 400 \\
Spatial Scale ($\lambda$) & 5.5 \\
Grid Influence Weight ($\eta$) & 0.35 \\
\hline
\end{tabular}
\end{center}

Table~\ref{tab:grid_params} summarizes the grid cell parameters used within the experiments. Grid cells were divided into 8 modules, with 50 cells per module, resulting in a total of 50 $\times$ 8 = 400 grid cells in the model. A spatial scale of $\lambda=5.5$ yields an activation diameter of roughly 2.7m. It is important to note that during the "no grid cell" experiments, $\eta$ is set to 0 to eliminate any effect of grid cells on the model.  

\newpage

\begin{center}
\captionof{table}{Place Cell Learning \& Inhibition Parameters}
\label{tab:pc_params}

\begin{tabular}{l c}
\hline
\textbf{Parameter} & \textbf{Value} \\
\hline
BVC Normalization Factor ($\alpha_{pb}$) & $\sqrt{0.4} \approx 0.632$ \\
GC Normalization Factor ($\alpha_{pg}$) & $\sqrt{0.4} \approx 0.632$ \\
Recurrent Inhibition Gain ($\Gamma^{pp}$) & $0.70$ \\
BVC Afferent Inhibition Gain ($\Gamma^{pb}$) & $0.35$ \\
GC Afferent Inhibition Gain ($\Gamma^{pg}$) & $0.35$ \\
Learning Time Constant ($\tau_w$) & 10 timesteps (960\,ms) \\
\hline
\end{tabular}
\end{center}

Table~\ref{tab:pc_params} summarizes the place cell learning and inhibition parameters. Both normalization factors were set to $\sqrt{0.4}$, and the afferent inhibitory gains for BVC and GC inputs were each set to half the recurrent inhibition gain.

\subsection{Metrics for Spatial Aliasing}
\label{sec:spa_aliasing}

Spatial aliasing was measured using the Spatial Aliasing Index (SAI) and Mean Spatial Aliasing Index (MSAI) metrics to examine how place cells responded in different regions of an environment. For a bin $i$ located at $(x_i, y_i)$, the SAI is given by:
\begin{equation}
\begin{aligned}
\mathrm{SAI}^{(i)} &=
\frac{1}{N}\sum_{\substack{j=1\\ j\neq i}}^{N}
\mathbf{1}\!\left\{\lVert (x_i,y_i)-(x_j,y_j)\rVert > d_{\mathrm{th}}\right\} \\
&\hspace{6em}\cdot \mathrm{CosSim}\!\left(\mathbf{a}^{(i)},\mathbf{a}^{(j)}\right)
\end{aligned}
\end{equation}

\noindent where:
\begin{itemize}
    \item $N$ is the total number of bins.
    \item $d_{\mathrm{th}}$ is the distance threshold to exclude nearby bins.
    \item $\mathbf{1}\{\cdot\}$ is the indicator function (equal to 1 if the distance between bins exceeds $d_{\mathrm{th}}$, and 0 otherwise).
    \item $\mathrm{CosSim}(\mathbf{a}^{(i)}, \mathbf{a}^{(j)})$ is the cosine similarity between the place cell activation vectors in bins $i$ and $j$.
\end{itemize}
To evaluate the model's overall aliasing performance, the MSAI is computed as the average of the SAI across all bins:
\begin{equation}
\mathrm{MSAI} = \frac{1}{N} \sum_{i=1}^{N} \mathrm{SAI}^{(i)}
\label{eq:msai}
\end{equation}
Larger values of $\mathrm{SAI}^{(i)}$ indicate greater spatial aliasing. This means that spatially distant bins exhibit similar place-cell activation patterns. The MSAI aggregates this effect across the environment, with higher MSAI values suggesting more aliasing. Conversely, lower $\mathrm{SAI}^{(i)}$ and MSAI values indicate stronger place-cell localization and improved spatial discrimination.

\section{Results \& Discussion}
\label{sec:results}

\begin{figure}[t]
    \centering
    \includegraphics[width=0.9\columnwidth]{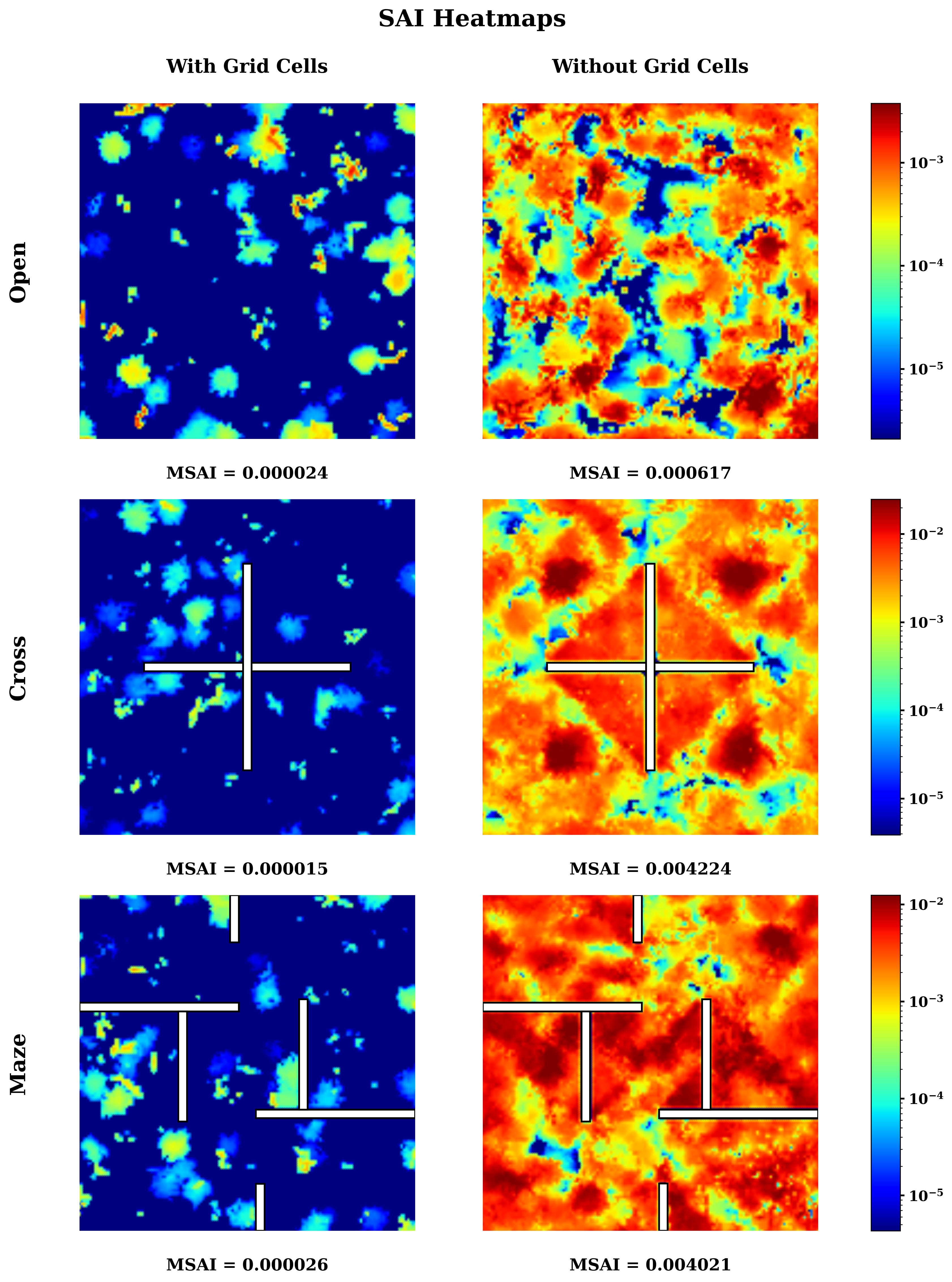}
    \caption{Spatial Aliasing Index (SAI) heatmaps with and without grid cells across environments. Rows show Open, Cross, and Maze (top to bottom); columns show with grid cells (left) and without grid cells (right). Color indicates aliasing magnitude on a logarithmic scale (lower is better). Reported Mean Spatial Aliasing Index (MSAI) values are shown beneath each heatmap.}
    \label{fig:msai_heatmaps}
\end{figure}

\begin{center}
\captionof{table}{Mean Spatial Aliasing Index (MSAI, $\times 10^{-4}$) across environments
with and without grid cells (GC). Values report mean $\pm$ standard deviation
over five independent trials.}
\label{tab:msai_summary}

\begin{tabular}{l c c c}
\hline
\textbf{Environment} & \textbf{GC} & \textbf{No GC} & \textbf{Improvement} \\
\hline
Open  & $0.3 \pm 0.1$ & $5.7 \pm 0.3$  & $94.73\%$ \\
Cross & $0.3 \pm 0.1$ & $42.4 \pm 1.0$ & $99.29\%$ \\
Maze  & $1.3 \pm 2.2$ & $38.9 \pm 2.7$ & $96.65\%$ \\
\hline
\end{tabular}
\end{center}
This section summarizes the results of the model's performance with and without grid cells in three different environments: Open, Cross, and Maze, as shown in Fig. \ref{fig:environments}. A total of five trials were conducted for each configuration (with and without grid cells) in each environment, and the MSAI scores were averaged for the five trials. 
The results are shown in Table~\ref{tab:msai_summary}. 

Examples of MSAI heatmaps from one trial for all three environments are shown in Fig.~\ref{fig:msai_heatmaps}. Consistent with the averaged MSAI values in Table~\ref{tab:msai_summary}, the heatmaps show lower SAI across most spatial bins when grid-cell input is included, with the most pronounced reduction in the Cross environment. These visualizations complement the results shown in Table~\ref{tab:msai_summary} by localizing where aliasing is mitigated within each environment.

An important detail to consider is how prone a particular environment is to aliasing when using place representations based purely on boundary vector cells (BVC). The ``Cross'' environment, for example, divides the map into four rotationally symmetric quadrants, as shown in Fig.~\ref{fig:environments}. This symmetry creates sensory similarity within the quadrants, increasing the potential for place-field aliasing. In models where place representations are constructed primarily from BVC inputs, such environments are therefore especially prone to aliasing~\cite{skaggs1998spatial}.

Adding a source of path integration signals to the model, namely grid cells, substantially improves the disambiguation of these perceptually similar regions, as evidenced by a reduction of approximately 99.3\% in the averaged Mean Spatial Aliasing Index (MSAI) in five trials in the Cross environment. 

The Open environment exhibits more moderate baseline aliasing due to fewer repeated boundary configurations. The Maze environment shows substantial but more variable improvement: although grid cells mitigate much of the BVC-induced aliasing, corridors introduce repeated spatial patterns, which can cause distant locations to retain moderate similarity in their place-cell representations, contributing to higher and more variable MSAI values.

\section{Conclusion \& Future Work}
\label{sec:conclusion}
Overall, our findings demonstrate that adding a source of path integration signals, namely grid cells, helps mitigate the degree of aliasing exhibited in place-cell representations across environments with varying degrees of structural symmetry. Future work will investigate the use of an attractor-based grid cell model and whether its biological plausibility introduces different trade-offs in aliasing mitigation and stability. In particular, it would be valuable to implement 3D attractor-based grid cells (accompanied by other relevant 3D cells, such as BVCs, place cells, and head direction cells) \cite{GerstenslagerDukenbaevMinai2025}, and examine the unique challenges that may arise by testing this implementation in varied, real-world environments. Such extensions would help clarify the challenges associated with scalable, biologically grounded spatial representations and move the model toward greater robustness in realistic navigation settings.

\section*{Acknowledgments}
The authors thank Andrew Gerstenslager and Bekarys Dukenbaev for helpful discussions and explanations of previous models. All results and analyses presented in this paper are solely the work of the authors.

\balance

\end{document}